\documentclass[conference]{IEEEtran}
\IEEEoverridecommandlockouts
\usepackage[pagebackref,breaklinks,colorlinks]{hyperref}
\usepackage{cite}
\usepackage{amsmath,amssymb,amsfonts}
\usepackage{graphicx}
\usepackage{textcomp}
\usepackage{caption}
\usepackage{url}
\usepackage{cleveref}
\usepackage{multirow}
\usepackage{algpseudocode}
\usepackage{algorithm}
\usepackage{xcolor} % Allows coloring of table cells
\usepackage{colortbl}
\usepackage{pifont}
\usepackage{multirow}
\usepackage{dblfloatfix}

\makeatletter
\renewcommand{\paragraph}{%
  \@startsection{paragraph}{4}%
  {\z@}{0.2em}{-1em}%
  {\normalfont\normalsize\bfseries}%
}
\makeatother

\definecolor{red}{rgb}{0.95,0.4,0.4}
\definecolor{green}{rgb}{0.55,1.0,0.55}
\definecolor{lightgreen}{rgb}{0.75,1.0,0.75}
\definecolor{blue}{rgb}{0.4,0.4,0.95}
\definecolor{darkblue}{rgb}{0,0,0.8}
\definecolor{darkred}{rgb}{0.8,0,0}
\definecolor{darkgreen}{rgb}{0,0.5,0}
\definecolor{grey}{rgb}{0.6,0.6,0.6}
\definecolor{amber}{RGB}{255,210,43}
\definecolor{lightgray}{rgb}{0.97,0.97,0.97}

\def\BibTeX{{\rm B\kern-.05em{\sc i\kern-.025em b}\kern-.08em
    T\kern-.1667em\lower.7ex\hbox{E}\kern-.125emX}}
\begin{document}
\pagestyle{plain} % Default style with page numbers at the bottom center
\pagenumbering{arabic} % Use Arabic numerals (1, 2, 3, ...)

\title{Structured Semantic 3D Reconstruction (S23DR) Challenge 2025 - Winning solution
}

\author{\IEEEauthorblockN{Jan Skvrna}
\IEEEauthorblockA{\textit{Dept. of Cybernetics} \\
\textit{FEE, Czech Technical University}\\
skvrnjan@fel.cvut.cz}
\and
\IEEEauthorblockN{Lukas Neumann}
\IEEEauthorblockA{\textit{Dept. of Cybernetics} \\
\textit{FEE, Czech Technical University}\\
lukas.neumann@cvut.cz}
}

\setlength{\abovedisplayskip}{8pt}
\setlength{\belowdisplayskip}{8pt}

\maketitle

\begin{abstract}

This paper presents the winning solution for the S23DR Challenge 2025, which involves predicting a house's 3D roof wireframe from a sparse point cloud and semantic segmentations. Our method operates directly in 3D, first identifying vertex candidates from the COLMAP point cloud using Gestalt segmentations. We then employ two PointNet-like models: one to refine and classify these candidates by analyzing local cubic patches, and a second to predict edges by processing the cylindrical regions connecting vertex pairs. This two-stage, 3D deep learning approach achieved a winning Hybrid Structure Score (HSS) of 0.43 on the private leaderboard.

\end{abstract}    
\section{Introduction}
\label{sec:intro}

This paper presents the winning solution for the Structured Semantic 3D Reconstruction (S23DR) Challenge 2025. The primary objective is to predict the 3D wireframe of a house roof, a significant challenge in geometric computer vision. The challenge utilizes the new HoHo25k dataset~\cite{S23DR_2025}, which comprises approximately 25,000 house instances from across the US.

For each house instance, the provided data includes a sparse point cloud and camera parameters generated by the Structure-from-Motion method COLMAP~\cite{schonberger2016structure}. These are supplemented with per-image metric depth maps from Metric3Dv2~\cite{hu2024metric3d} and two types of semantic segmentations (ADE20k~\cite{zhou2017scene} and Gestalt). While the original images are not shared due to privacy concerns, these derived data products form the basis for reconstruction. The goal is to reconstruct the ground-truth wireframe, which is defined by a set of 3D vertices and the edges connecting them.

Performance is evaluated using the Hybrid Structure Score (HSS), a novel metric designed for this challenge. The HSS is computed as the harmonic mean of the vertex F1 score and the edge Intersection over Union (IoU):
\begin{multline}
\label{eq:hss}
\text{HSS}(gt_v, gte, pred_v, pred_e) = \\ \text{harmonic mean}(F1(gt_v, pred_v), \text{IoU}(gt_e, pred_e))
\end{multline}
where $gt_v$ and $pred_v$ are the ground-truth and predicted vertices, and $gt_e$ and $pred_e$ are the ground-truth and predicted edges. Notably, the HSS metric does not consider the semantic classification of vertices or edges, focusing solely on their geometric accuracy.

Our proposed solution departs significantly from the baseline's 2D-centric approach. Instead of detecting features in 2D images and lifting them to 3D, our pipeline operates directly on the 3D point cloud. We first identify vertex candidates by leveraging the provided 2D segmentations to cluster points in 3D. Subsequently, we employ a PointNet-like architecture~\cite{qi2017pointnet} to refine the 3D coordinates of these vertices and a second, similar model to predict the edges connecting them. This direct 3D methodology proved highly effective, substantially boosting the final score while maintaining low computational overhead, ultimately securing the winning position in the challenge.

\section{Dataset understanding}
\label{sec:intro}

Before detailing our method, it is crucial to analyze the dataset's composition and inherent challenges, as they heavily influence our architectural decisions. The inputs for this task are more complex than typical computer vision problems, requiring careful consideration.

\begin{figure}
    \centering
    \includegraphics[width=0.49\linewidth]{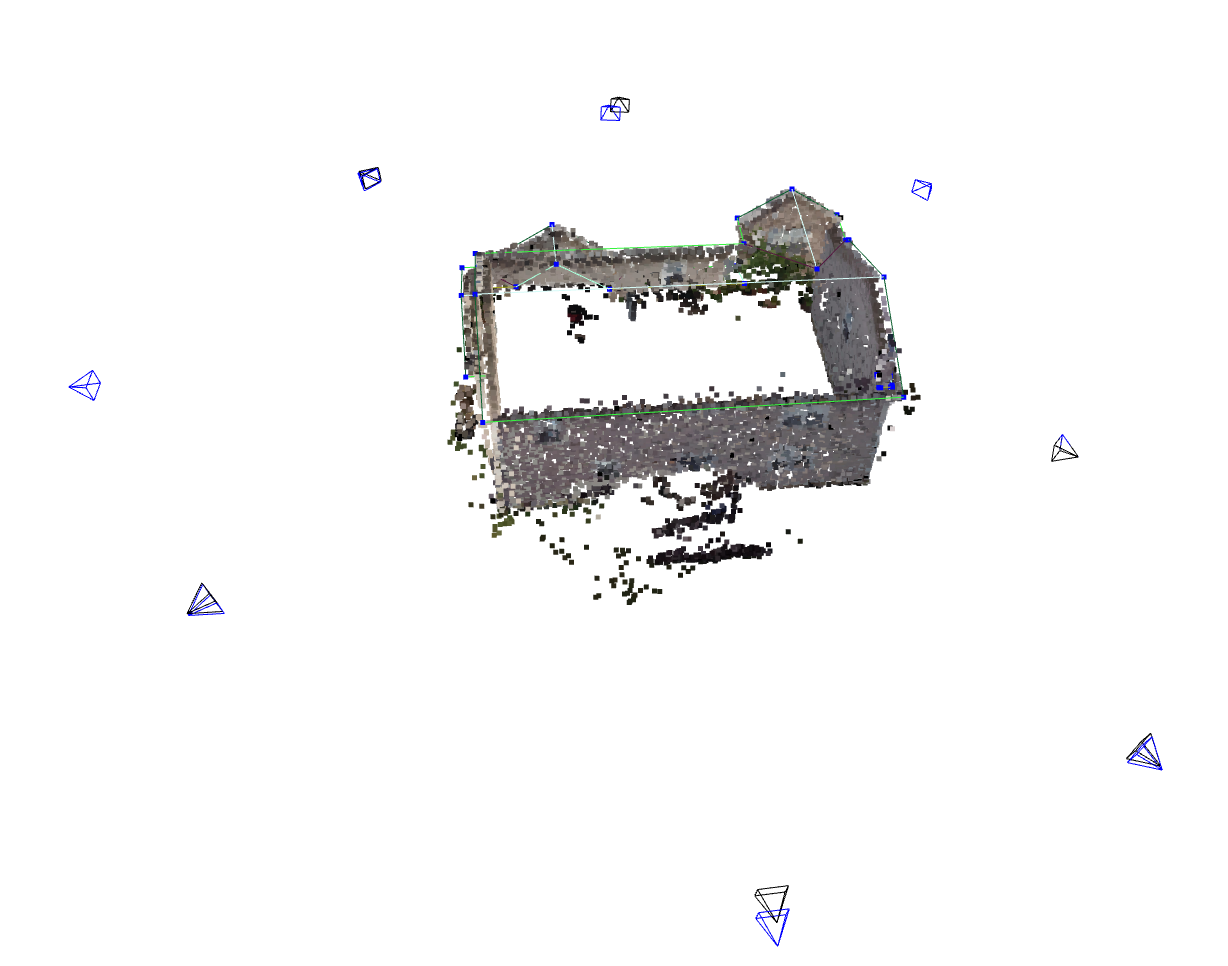}
    \includegraphics[width=0.49\linewidth]{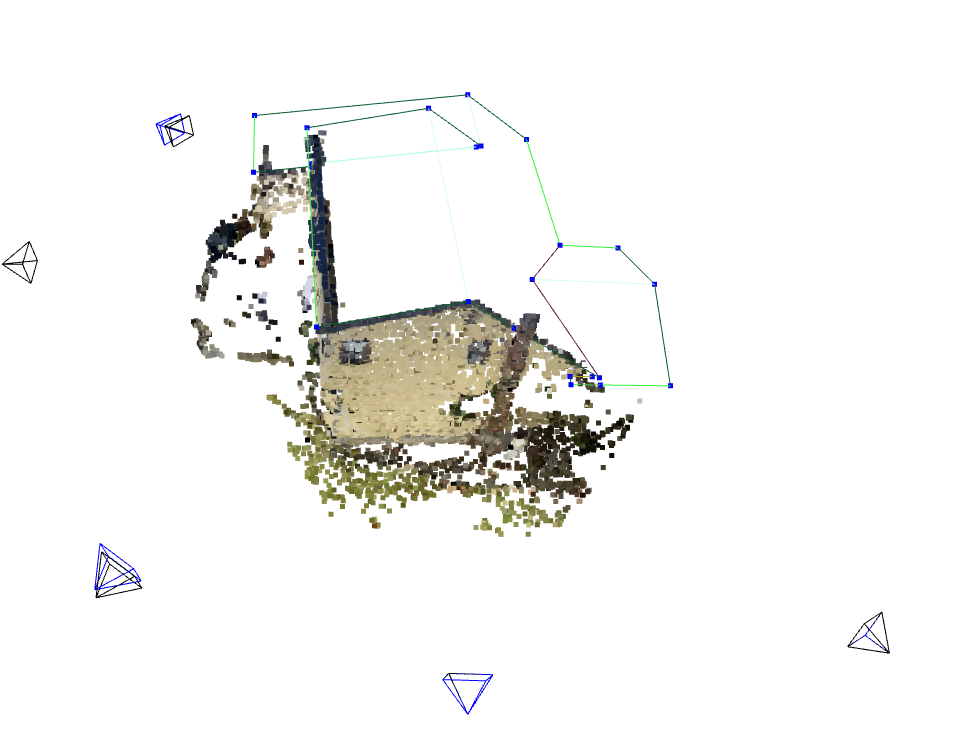}
    \caption{Two instances in the training set showing how the COLMAP~\cite{schonberger2016structure} point cloud, camera poses, and ground truth look like. Blue points and green lines represent ground truth vertices and edges.}
    \label{fig:frames}
\end{figure}

The dataset was generated by processing multiple images of each building. For each image, a metric depth map was computed using Metric3Dv2~\cite{hu2024metric3d}, and two different semantic segmentations (ADE20k and Gestalt) were predicted. The Structure-from-Motion method COLMAP~\cite{schonberger2016structure} was then used on the set of (unreleased) images to generate a sparse 3D point cloud, along with camera poses and intrinsic parameters for each view. Figure~\ref{fig:frames} provides two examples from the training set, illustrating the relationship between the point cloud, camera poses, and the ground-truth wireframe.

From our analysis, we identified two primary challenges that need to be taken into account:

\subsubsection{Incomplete Point Clouds}

A significant challenge is the frequent incompleteness of the COLMAP point clouds. As shown in Figure~\ref{fig:frames}, the reconstruction often fails to capture entire sections of a house, either due to occlusions, lack of texture, or insufficient image coverage. Despite this missing geometric data, the task requires predicting the \textit{full} house wireframe. This necessitates a model that can infer and complete unseen structures, making the problem substantially harder than simple geometry fitting.

\subsubsection{Inconsistent Camera Parameters and Data Misalignment}

\begin{figure}
    \centering
    \includegraphics[width=0.49\linewidth]{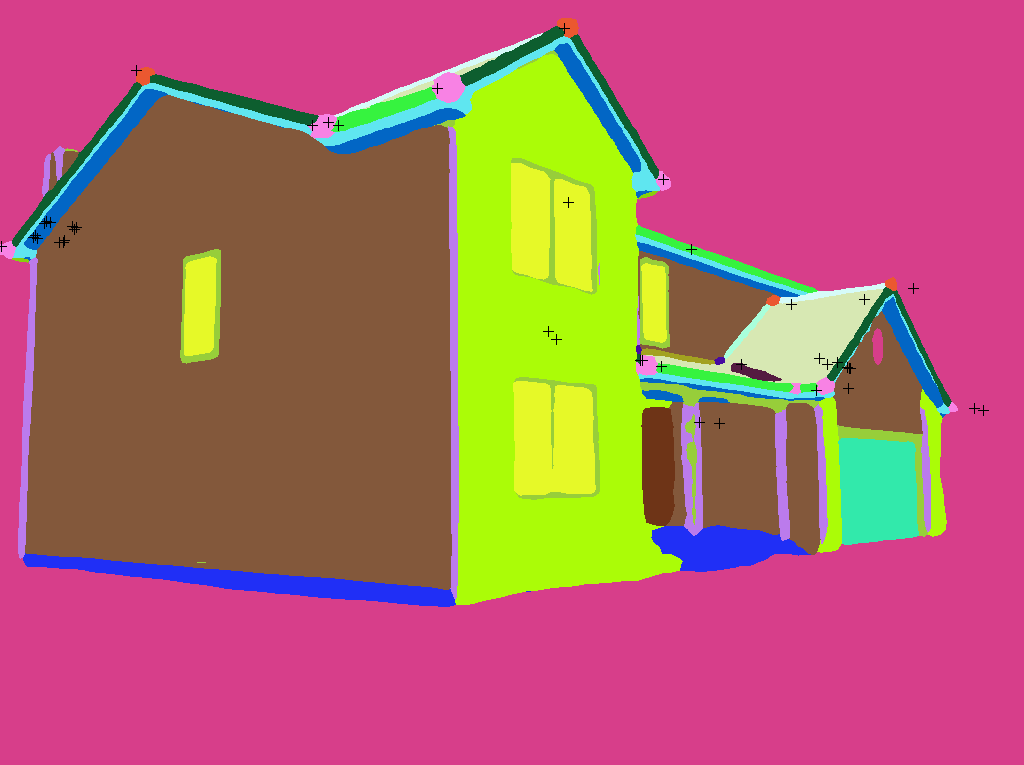}
    \includegraphics[width=0.49\linewidth]{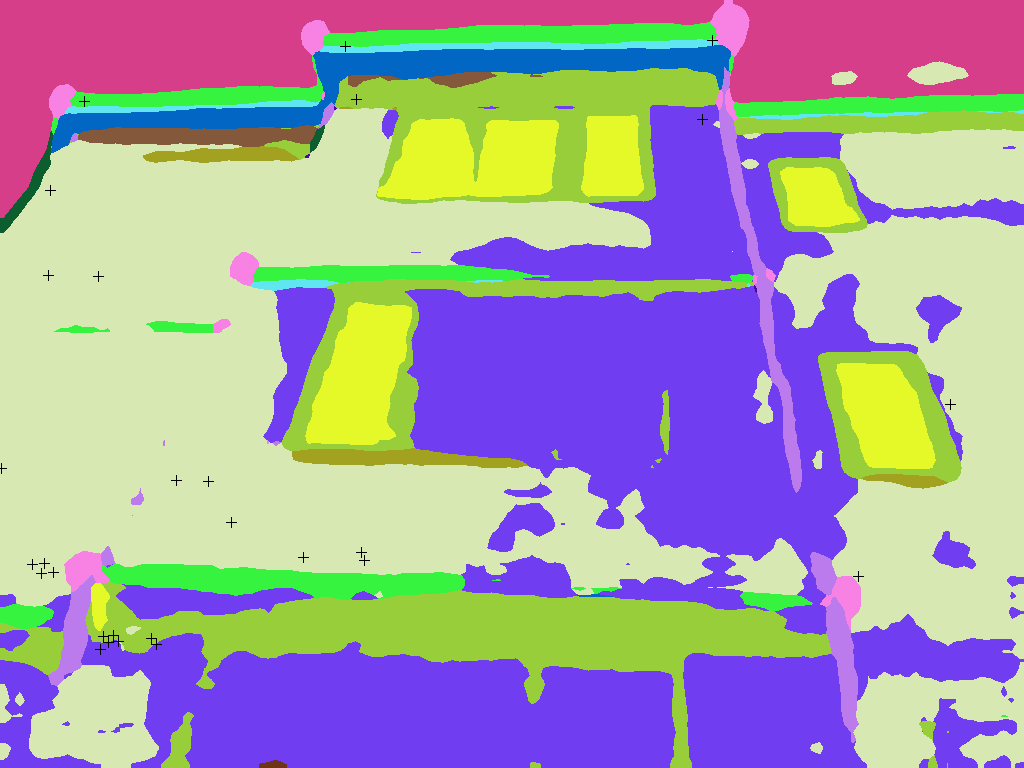}
    \caption{Gestalt segmentations of two different houses, where black crosses depict projected 3D ground truth vertices given the estimated $K, R, t$.}
    \label{fig:gestalt}
\end{figure}

A more subtle but critical issue is the inconsistency in the provided camera parameters ($K, R, t$). For reasons not specified in the dataset documentation, there appear to be two distinct sets of camera parameters, which sometimes differ from one another (visualized as black vs. blue cameras in Figure~\ref{fig:frames}). This discrepancy causes a misalignment between the 3D data (the point cloud) and the 2D data (the segmentations). As illustrated in Figure~\ref{fig:gestalt}, when ground-truth 3D vertices are projected using the provided camera parameters, they often do not align perfectly with the corresponding features in the Gestalt segmentation. This misalignment complicates any approach that relies on fusing information from both the 3D point cloud and the 2D segmentations.

\section{Method}
\label{sec:intro}

Our method represents a fundamental shift from the baseline, which relies on a fragile 2D-to-3D lifting process. We instead propose a robust pipeline that operates directly in 3D, using deep learning to first refine vertex geometry and then predict their connectivity. This section first summarizes the baseline to highlight its limitations, then details our multi-stage solution.

\subsection{Baseline Solution and Motivation}
The baseline method iterates through each image, detecting 2D vertex candidates by clustering pixels in the Gestalt segmentation (e.g., `apex`). Edges are subsequently inferred by fitting lines to other segmentation classes (e.g., `ridge`). The critical step is lifting these 2D detections into 3D by finding a corresponding point in the COLMAP~\cite{schonberger2016structure} point cloud or, as a fallback, using the metric depth map. The resulting 3D points are then merged and filtered.

This approach is highly susceptible to errors in the initial 2D detection and, more critically, the 2D-to-3D lifting. As discussed in Section II, the frequent misalignments between the 2D segmentations and the 3D point cloud make this lifting process inherently inaccurate. These weaknesses served as the primary motivation for developing our 3D-centric approach.

\subsection{Proposed Pipeline}
Our method replaces this brittle process with a robust, three-stage pipeline that operates entirely in the 3D domain.
\begin{enumerate}
    \item \textit{3D Vertex Candidate Generation:} We first generate coarse 3D vertex candidates by identifying and merging point clusters from the COLMAP cloud that correspond to vertex-like regions in the Gestalt segmentations across all views.
    \item \textit{Vertex Refinement and Classification:} We use a multi-task PointNet-like~\cite{qi2017pointnet} model to process a local 3D patch around each candidate. This model refines the candidate's 3D position, classifies it as a valid vertex, and predicts a confidence score.
    \item \textit{Edge Prediction:} Using a second PointNet-like model, we analyze the 3D region between pairs of the refined vertices to predict the existence of an edge.
\end{enumerate}

\subsubsection{Stage 1: 3D Vertex Candidate Generation}
To bypass the 2D-to-3D lifting, we generate candidates directly in 3D. We iterate through all views and identify clusters of relevant classes in the Gestalt segmentation (`apex`, `eave end point`, `flashing end`). To mitigate the misalignment shown in~\cref{fig:gestalt}, we apply iterative binary dilation to each cluster mask, expanding it until it captures at least five points from the COLMAP cloud. This ensures we have sufficient local geometry for each potential detection.

\begin{algorithm}
\caption{Extended PointNet Architecture}
\begin{algorithmic}[1]
\State \textbf{Input:} $x \in \mathbb{R}^{B \times 11 \times N}$ \Comment{$B$: batch size, $N \leq 1024$ points}
\Statex
\State \textit{// Point-wise Feature Extraction}
\State Conv1D($11 \to 64$), Conv1D($64 \to 128$), Conv1D($128 \to 256$),
\State Conv1D($256 \to 512$), Conv1D($512 \to 1024$), Conv1D($1024 \to 1024$), Conv1D($1024 \to 2048$)
\State \textit{// Each followed by BatchNorm1D + LeakyReLU, residual at 6th convolution}
\Statex
\State \textit{// Channel Attention}
\State $\text{pool} \gets \mathrm{AdaptiveAvgPool1D}(x_7)$
\State $\alpha \gets \mathrm{Conv1D}(2048 \to 128, 1) \to \mathrm{ReLU} \to \mathrm{Conv1D}(128 \to 2048, 1) \to \mathrm{Sigmoid}(\text{pool})$
\State $x_{\mathrm{att}} \gets x_7 \odot \alpha$ \Comment{Element-wise multiplication}
\Statex
\State \textit{// Multi-scale Global Pooling}
\State $g_{\max} \gets \max_{i=1}^{N} x_{\mathrm{att},i}$
\State $g_{\mathrm{avg}} \gets \frac{1}{N}\sum_{i=1}^{N} x_{\mathrm{att},i}$
\State $g \gets 0.7 \cdot g_{\max} + 0.3 \cdot g_{\mathrm{avg}} \in \mathbb{R}^{B \times 2048}$
\Statex
\State \textit{// Shared Feature Extraction}
\State Linear($2048 \to 1024$), Linear($1024 \to 512$), Linear($512 \to 512$)
\State \textit{// GroupNorm + LeakyReLU + Dropout, residual at last shared layer}
\Statex
\State \textit{// Multi-head Prediction}
\State \textbf{Position:} Linear($512 \to 512 \to 256 \to 128 \to 64 \to 3$)
\State \textbf{Score:} Linear($512 \to 512 \to 256 \to 128 \to 64 \to 1$), Softplus
\State \textbf{Class:} Linear($512 \to 512 \to 256 \to 128 \to 64 \to 1$)
\State \textit{// Each with LeakyReLU + Dropout between layers}
\Statex
\State \textbf{Output:} 3D offset, confidence score, classification logits
\label{algo:pnet}
\end{algorithmic}
\end{algorithm}

After generating these point clusters from all views, we merge them. We first filter the global point cloud to retain only points belonging to at least one cluster. For each cluster, we compute its centroid and define it by all points within a 0.5-meter radius. Finally, any two clusters sharing over 50\% of their points are merged into a single candidate, yielding a final set of coarse 3D vertex candidates.

\subsubsection{Stage 2: Vertex Refinement and Classification}
The candidates from Stage 1 are coarse and require precise localization. For this, we designed a multi-task network, detailed in~\cref{algo:pnet}, to jointly classify and refine each candidate.

\textit{Input Patch Representation.} For each candidate, we extract a 4x4x4 meter cubic patch from the point cloud, centered at the candidate's centroid. Each point within this patch is represented by an 11-dimensional feature vector:
\begin{itemize}
    \item \textit{Relative Position (3D):} $(x,y,z)$ coordinates relative to the patch center.
    \item \textit{Color (3D):} Normalized RGB values $[-1, 1]$ from COLMAP.
    \item \textit{House Segmentation (1D):} A binary indicator from the ADE20k segmentation (`house` class).
    \item \textit{Gestalt Class (3D):} The Gestalt segmentation label, encoded as normalized RGB.
    \item \textit{Candidate Flag (1D):} A binary flag marking if the point was part of the original candidate cluster.
\end{itemize}

\textit{Network Architecture.} Our vertex network, described in~\cref{algo:pnet}, is an extended PointNet~\cite{qi2017pointnet} architecture. It employs point-wise convolutions to extract high-dimensional features, followed by channel attention and multi-scale global pooling to create a robust patch-level feature vector. This vector is then fed into three heads for multi-task learning:
\begin{enumerate}
    \item \textit{Classification Head:} Predicts if the patch contains a true vertex (supervised with Binary Cross-Entropy loss).
    \item \textit{Position Head:} Predicts a 3D offset from the patch center to the precise vertex location (Smooth L1 loss).
    \item \textit{Score Head:} Predicts a confidence score for the position offset, supervised by the L2 distance to the ground-truth offset (Smooth L1 loss).
\end{enumerate}
Following this stage, we discard candidates with low classification scores and apply the predicted offsets to the remaining ones, resulting in a set of refined 3D vertices.

\subsubsection{Stage 3: Edge Prediction}
With a refined set of vertices, the final task is to determine the edges connecting them. We frame this as a binary classification problem for each potential vertex pair.

\textit{Input Patch Representation.} For each pair of vertices, we define a cylindrical patch with a 1-meter radius around the line segment connecting them, extended by 1 meter at both ends to provide context. All points from the COLMAP cloud inside this cylinder form the input. Each point is represented by a 6D feature vector: relative $(x,y,z)$ position and normalized RGB color.

\textit{Network Architecture.} We use a second, simpler PointNet-style classifier, detailed in~\cref{algo:pnet_class}. It processes the cylindrical patch, applies global max-pooling to obtain a patch-level feature vector, and passes it through an MLP to classify the pair as an edge or not. For training, we construct a balanced dataset of positive and negative pairs to handle the natural class imbalance. After classifying each pair, we obtain our final wireframe.

\begin{algorithm}
\caption{Classification PointNet Architecture}
\begin{algorithmic}[1]
\State \textbf{Input:} $x \in \mathbb{R}^{B \times 6 \times N}$ where $B$ is batch size, $N \leq 1024$ points
\Statex
\State \textit{// Point-wise Feature Extraction}
\State Conv1D($6 \to 64$), Conv1D($64 \to 128$), Conv1D($128 \to 256$),
\State Conv1D($256 \to 512$), Conv1D($512 \to 1024$), Conv1D($1024 \to 2048$)
\State \textit{// Each followed by BatchNorm1D + ReLU}
\Statex
\State \textit{// Global Aggregation}
\State $g \gets \max_{i=1}^{N} x_i \in \mathbb{R}^{B \times 2048}$ \Comment{Max pooling over point dimension}
\Statex
\State \textit{// Classification MLP}
\State Linear($2048 \to 1024$), Linear($1024 \to 512$), Linear($512 \to 256$),
\State Linear($256 \to 128$), Linear($128 \to 64$), Linear($64 \to 1$)
\State \textit{// ReLU + Dropout($p \in [0.3, 0.5]$) between layers, except final}
\Statex
\State \textbf{Output:} Binary classification logits $\in \mathbb{R}^{B \times 1}$
\label{algo:pnet_class}
\end{algorithmic}
\end{algorithm}

\subsection{Implementation and Training Details}
Both models are relatively lightweight and were trained on a single NVIDIA A100 GPU in a few hours using the AdamW optimizer with a batch size of 128. 

The vertex classification dataset we generated contains approximately 600,000 candidate samples from the 25,000 buildings in the training set. A similarly sized dataset was created for training the edge prediction model. We found this scale sufficient for training our models effectively. Due to time constraints, no data augmentation was employed, though it would likely help mitigate overfitting and further improve performance.

At inference time, the process is efficient. On an NVIDIA T500 GPU, a single patch inference takes under 10 ms. While the number of vertex candidates is typically low (around 50 per house), the edge prediction model must run on all pairs, making it the more computationally intensive step.

\begin{figure}
    \centering
    \includegraphics[width=0.49\linewidth, height=4cm, keepaspectratio]{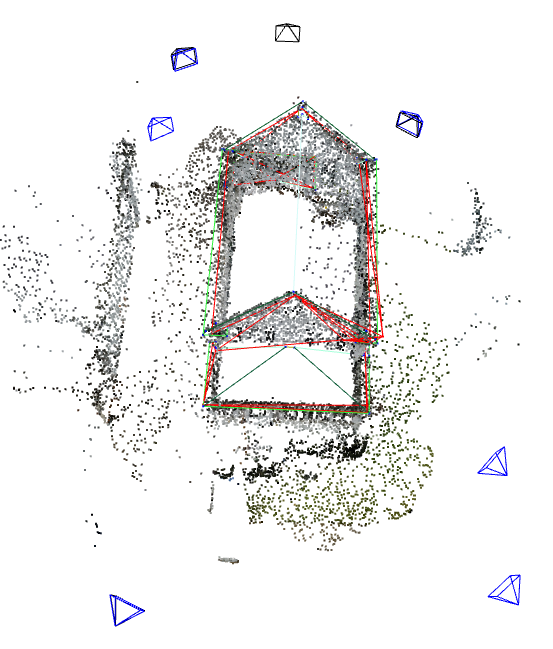}
    \includegraphics[width=0.49\linewidth]{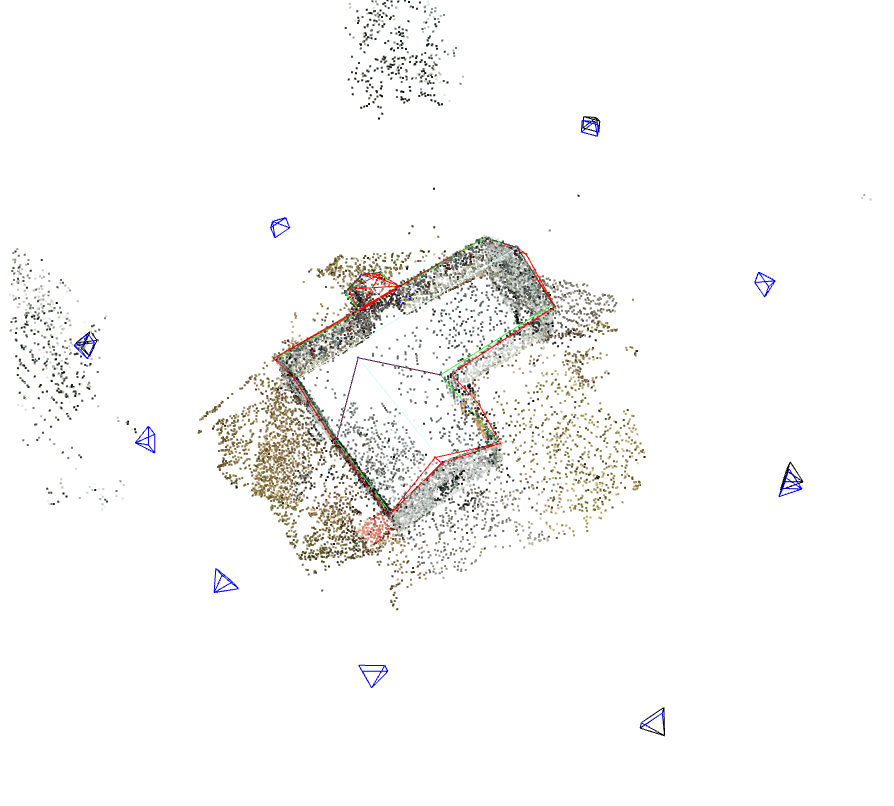}
    \includegraphics[width=0.49\linewidth]{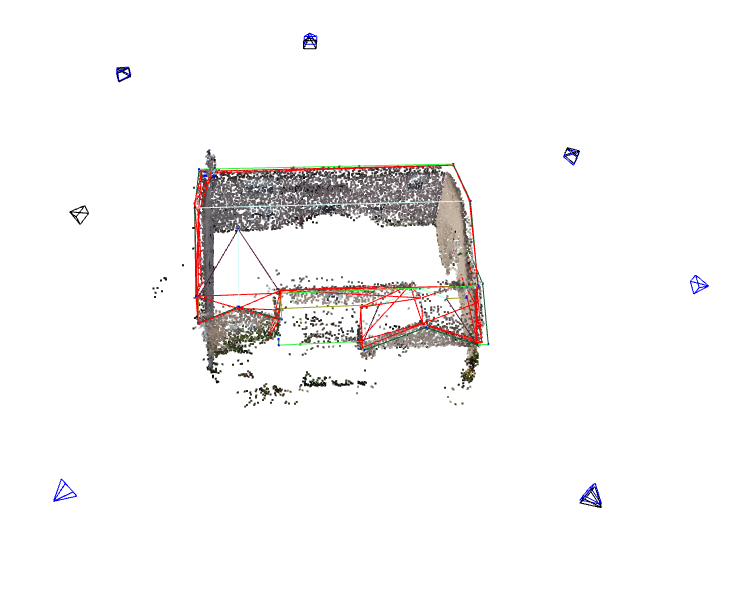}
    \includegraphics[width=0.49\linewidth]{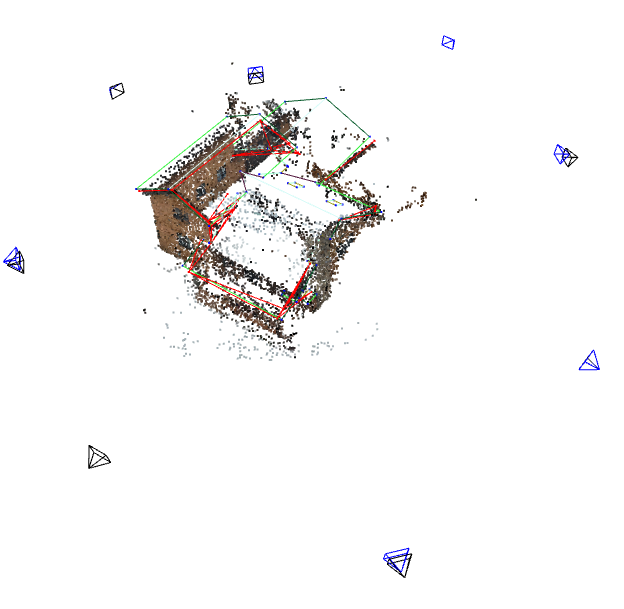}
    \caption{Qualitative analysis of four instances of building using our method. Blue vertices and green lines represent ground truth vertices and edges, while red vertices and lines represent predictions by our method.}
    \label{fig:qual}
\end{figure}

We also include a few instances with predicted wireframe as qualitative analysis in ~\cref{fig:qual}. 
\section{Experiments}
\label{sec:conclusion}

\begin{table}[]
    \centering
    \begin{tabular}{| c | c | c | c |}
    \hline  
         & Mean HSS & Mean F1 & Mean IoU \\
         \hline 
        Baseline & 0.148 & 0.220 & 0.122 \\
        + tuned parameters & 0.233 & 0.304 & 0.200 \\
        \hline
        Our method & 0.257 & 0.345 & 0.217 \\
        + vertex classification & 0.260 & 0.387 & 0.207 \\
        + edge classification & 0.286 & 0.387 & 0.239 \\
        \hline
    \end{tabular}
    \caption{Ablation study evaluating multiple adjustments made to the baseline method on the validation set.}
    \label{tab:ablation}
\end{table}
We provide shallow ablation studies for this method. First, the~\cref{tab:ablation} shows how each adjustment to the baseline improves the performance on the validation set. Tuning the hyperparameters for the baseline method significantly improves the performance.

Our method, without any filtering by vertex and edge classification, improves the performance by a small margin. Instead of lifting 2D centres into 3D, we segment all points in 3D and then take the centre of the points. An edge detection is done in the same manner as the baseline method. This small adjustment shows a small improvement. However, it produces noisy outputs. To tackle this, we first add the vertex classifier, which improves the mean F1 significantly, but also decreases the mean IoU. By adding the edge classifier, we further improve the mean IoU, which allows us to achieve the winning performance.

\begin{figure}[!h]
    \centering
    \includegraphics[width=1.\linewidth]{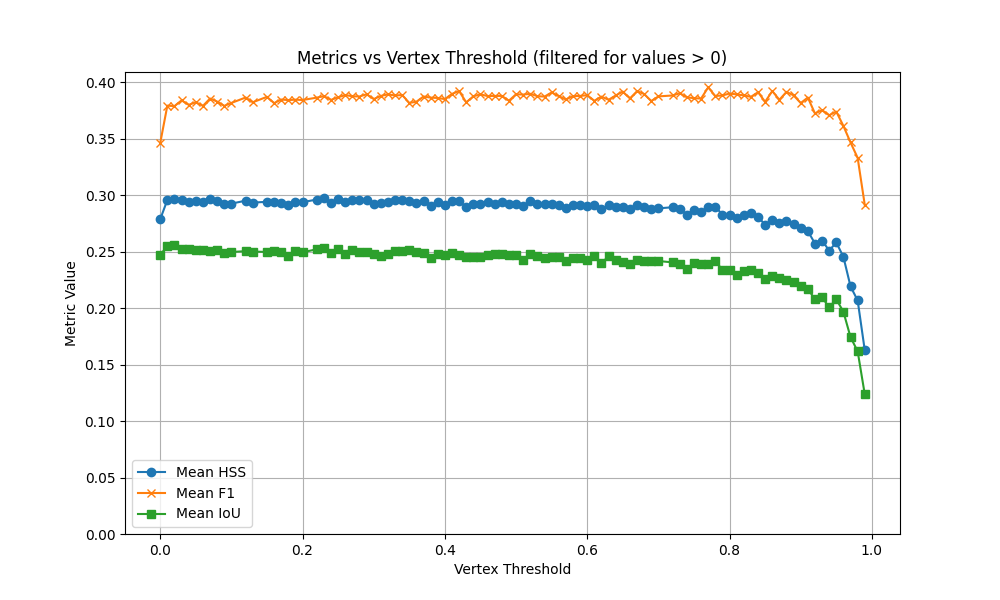}
    \caption{Ablation study on how the vertex classifier threshold affects the performance}
    \label{fig:vertex_threshold}
\end{figure}

In~\cref{fig:vertex_threshold} we provide an ablation how the choice of the vertex classifier threshold affects the overall HSS, F1 and IoU. The results show, that the threshold choice is not crucial if the threshold is under 0.7.

\begin{figure}[!h]
    \centering
    \includegraphics[width=1.\linewidth]{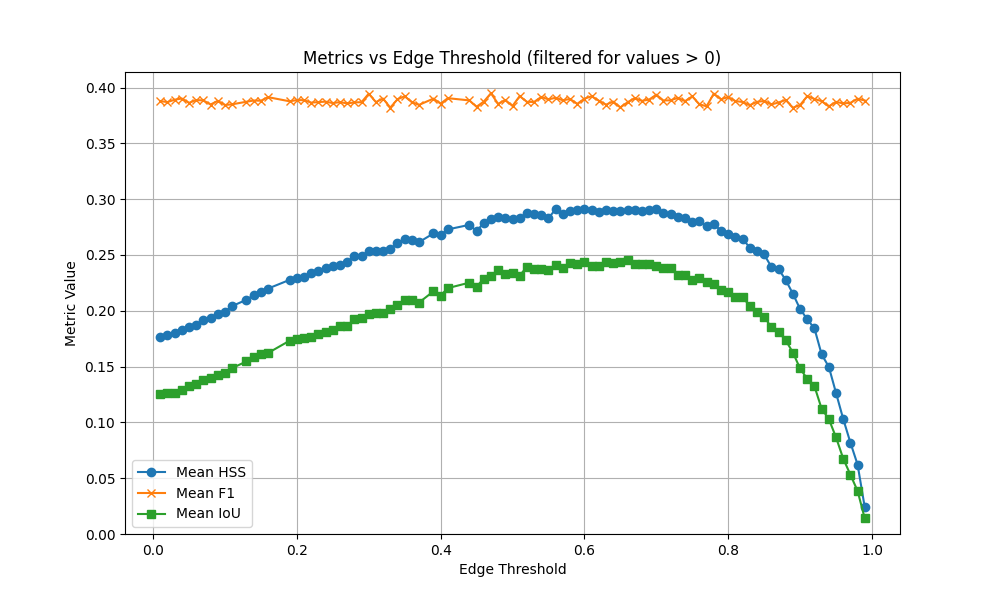}
    \caption{Ablation study on how the vertex classifier threshold affects the performance}
    \label{fig:edge_treshold}
\end{figure}

In~\cref{fig:edge_treshold} we provide an ablation how the choice of the edge classifier threshold affects the overall HSS, F1 and IoU. The results show that the choice of the threshold significantly affects the performance, showing that fine evaluation of the threshold is needed to achieve the best performance. As expected, the F1 is not affected by the choice of the threshold.

The experiments conducted show that the best thresholds are 0.59 for the vertex classifier threshold and 0.65 for the edge classifier threshold.

\section{Conclusion}
\label{sec:conclusion}

This paper presented the winning solution for the S23DR 2025 Challenge, achieving a mean Hybrid Structure Score (HSS) of 0.43 on the private leaderboard. The core of our success lies in abandoning the brittle 2D-to-3D lifting strategy of the baseline in favor of a direct 3D pipeline. By employing two specialized PointNet-like~\cite{qi2017pointnet} networks—one for refining vertex candidates and another for predicting their connectivity—we demonstrated that a targeted deep learning approach can yield significant performance gains while maintaining low inference times. The large scale of the dataset, providing over 600,000 training samples for our models, proved sufficient for training these relatively simple architectures effectively.

Despite its strong performance, our method's primary limitation is its dependency on the quality of the upstream input data: the COLMAP~\cite{schonberger2016structure} point cloud and the Gestalt segmentations. Errors in these inputs, such as missing geometry in the point cloud or misalignments in the segmentations, directly impact our pipeline's ability to generate accurate candidates and can lead to reconstruction failures from which our models cannot recover. Conversely, this dependency suggests that the performance of our framework would naturally improve with access to higher-fidelity input data, such as denser point clouds or more precise semantic labels.

Looking forward, several ideas exist for future improvement. First, employing more advanced point-based architectures (e.g., those incorporating attention or more sophisticated local feature aggregators) could capture more complex geometric relationships and improve robustness. Second, the integration of robust 3D data augmentation, including random rotations, scaling, noise injection, and point dropout, would almost certainly enhance model generalization and mitigate the overfitting. Finally, exploring a joint-learning framework that predicts vertices and edges in a more integrated, or even end-to-end, fashion could lead to more globally consistent wireframes. Overall, our work validates that targeted deep learning on 3D point sets is a powerful and efficient paradigm for structured 3D reconstruction tasks.

\section*{Acknowledgement}
The research was supported by Czech Science Foundation Grant No. 24-10738M. The access to the computational infrastructure of the OP VVV funded project CZ.02.1.01/0.0/0.0/16\_019/0000765 ``Research Center for Informatics'' is also gratefully acknowledged. We also acknowledge the support from the Student Grant Competition of the Czech Technical University in Prague, grant No. SGS23/173/OHK3/3T/13.

\bibliographystyle{ieeetr}
\bibliography{main}

\end{document}